\documentclass[10pt,twocolumn,letterpaper]{article}

\usepackage{wacv}
\usepackage{times}
\usepackage{epsfig}
\usepackage{graphicx}

\usepackage{amsmath,amssymb,amsfonts}
\usepackage{algorithmic}
\usepackage{graphicx}
\usepackage{textcomp}

\usepackage{comment}
\usepackage{amsmath,amssymb} 
\usepackage{color}
\usepackage[T1]{fontenc}

\usepackage{multirow}
\usepackage{float}

\wacvfinalcopy 
\ifwacvfinal
\usepackage[breaklinks=true,bookmarks=false]{hyperref}
\else
\usepackage[pagebackref=true,breaklinks=true,colorlinks,bookmarks=false]{hyperref}
\fi

\makeatletter
\newcommand{\printfnsymbol}[1]{%
  \textsuperscript{\@fnsymbol{#1}}%
}

\begin{document}

\title{Noise as a Resource for Learning in Knowledge Distillation}

\author{Elahe Arani\thanks{Equal contribution.}, Fahad Sarfraz\printfnsymbol{1}, and Bahram Zonooz\\
Advanced Research Lab, NavInfo Europe, Eindhoven, The Netherlands\\
{\tt\small \{elahe.arani, fahad.sarfraz\}@navinfo.eu, bahram.zonooz@gmail.com}
}

\maketitle

\begin{abstract}
While noise is commonly considered a nuisance in computing systems, a number of studies in neuroscience have shown several benefits of noise in the nervous system from enabling the brain to carry out computations such as probabilistic inference as well as carrying additional information about the stimuli. Similarly, noise has been shown to improve the performance of deep neural networks. In this study, we further investigate the effect of adding noise in the knowledge distillation framework because of its resemblance to collaborative subnetworks in the brain regions. We empirically show that injecting constructive noise at different levels in the collaborative learning framework enables us to train the model effectively and distill desirable characteristics in the student model. In doing so, we propose three different methods that target the common challenges in deep neural networks: minimizing the performance gap between a compact model and large model (Fickle Teacher), training high performance compact adversarially robust models (Soft Randomization), and training models efficiently under label noise (Messy Collaboration). Our findings motivate further study in the role of noise as a resource for learning in a collaborative learning framework.

\end{abstract}

\section{Introduction}
Noise permeates every level of the nervous system, from the perception of sensory signals to the generation of motor responses~\cite{faisal2008noise}. Despite its pervasiveness, noise has been predominantly viewed as a nuisance in computing systems~\cite{maass2014noise}. Recently, though, there has been a shift in neuroscience and a number of studies argue for the beneficial role of noise \cite{faisal2008noise,maass2014noise,mcdonnell2011benefits}. Multiple noise sources contribute to trial-to-trial variability, variations in neural responses to the same stimuli, which can be considerably different across stimuli, suggesting that it could also provide an important contribution to the information conveyed by the neural responses about the stimuli~\cite{scaglione2011trial}. Instead of a mere consequence of inherently stochastic processes on the molecular level, noise and trial-to-trial variability can be considered as salient components of the computational strategy of the brain~\cite{maass2014noise}. The brain exploits noise to carry out computations, such as probabilistic inference through sampling~\cite{maass2014noise}. Furthermore, neural networks that have been formed in the presence of noise will be more robust and explore more states, which will facilitate learning and adaptation to the changing demands of a dynamic environment~\cite{faisal2008noise}.

Analogously, numerous empirical studies have shown that noise plays a crucial role in effective and efficient training of neural networks~\cite{zhou2019towards}. Noise has been used as a common regularization technique to improve the generalization performance of overparameterized deep neural networks (DNNs) by adding it to the input data, the weights, or the hidden units \cite{an1996effects,blundell2015weight,graves2011practical,srivastava2014dropout,steijvers1996recurrent,wan2013regularization}.
Many noise techniques have been shown to improve generalization such as Dropout \cite{srivastava2014dropout} and injection of noise to the gradients \cite{bottou1991stochastic,neelakantan2015adding}. Previous studies also showed that noise is crucial for non-convex optimization \cite{kleinberg2018alternative,li2017convergence,yim2017gift,zhou2017stochastic}. Zhou \etal~\cite{zhou2019towards} showed that noise enables the gradient descent algorithm to efficiently escape from the spurious local optimum and converge to a global optimum.

All of these studies suggest that noise can indeed be a critical resource for learning.
To further investigate the role of noise, we focus on the knowledge distillation framework because of its resemblance to the collaborative learning between different regions in the brain. It also enables training high-performance compact models for efficient real-world deployment on resource-constrained devices.
Knowledge distillation involves training a smaller model (student) under the supervision of a larger pre-trained model (teacher) and consistently provides generalization gains.
Despite the promising performance gain, there is still a significant generalization gap between the student and teacher. Consequently, an optimal method of capturing knowledge from the larger model and transferring it to a smaller model remains an open question.
Inspired by trial-to-trial variability in the brain, we introduce variability through noise at the input level, supervision signal from the teacher, or target level.
We propose novel ways of injecting noise into the knowledge distillation framework as general and scalable techniques and exhaustively evaluate their performance.

Our contributions are as follows:

\begin{itemize}
\item We empirically show that noise can be used as a critical resource for learning in the knowledge distillation framework and evaluate the effect of injecting constructive noise at different levels of the framework. 

\item ``Fickle Teacher" (FT), a novel approach to simulate trial-to-trial response variability of biological neural networks.
The method exposes the student to the uncertainty of the teacher which results in consistent generalization improvement over the vanilla knowledge distillation method (from 94.28\% to 94.67\%).

\item ``Soft Randomization" (SR), a novel approach for increasing robustness to input variability. The method considerably increases the capacity of the model to learn robust features with even small additive noise with a minimal loss in generalization compared to Gaussian data augmentation (GA). On SVHN, SR achieves 93.39\%  generalization with 51.39\% robustness compared to GA’s 93.22\% generalization and 25.94\% robustness for $\sigma=0.2$.

\item ``Messy Collaboration" (MC), an approach for using target variability as a strong deterrent to cognitive bias. We show the effectiveness of MC in learning with noisy labels.
\end{itemize}

\section{Methodology}

In this section, we provide details for the methods relevant to our study. 

\subsection{Knowledge Distillation.}
Hinton \etal \cite{hinton2015distilling} proposed to use the final softmax function with a raised temperature and use the smooth logits of the teacher as soft targets for the student. The method involves minimizing the Kullback–Leibler divergence between the smoother output probabilities:
\begin{equation}\label{equ_KD}
 \mathcal{L}=(1-\alpha)\mathcal{L}_{CE}(S(x),y)+\alpha\tau^2D_{KL}(S^\tau(x)||T^\tau(x))
\end{equation}
where $\mathcal{L}_{CE}$ denotes cross-entropy loss, $\tau$ and $\alpha$ are the hyperparameters which denote softmax temperature and balancing ratio. $S(.)$ denotes the softmax output of student, $S^\tau(.)$ and $T^\tau(.)$ denote the student's and teacher's softmax output with raised temperature, respectively.

\subsection{Out-of-Distribution Generalization}
Neural networks tend to generalize well when the test data comes from the same distribution as the training data \cite{deng2009imagenet,he2015deep}. 
However, models in the real world often have to deal with some form of domain shift which adversely affects the generalization performance of the models \cite{kawaguchi2017generalization,liang2017enhancing,moreno2012unifying,shimodaira2000improving}. 
Therefore, test set performance alone is not the optimal metric for evaluating the generalization of the models in the test environment. 
To measure the out-of-distribution performance, we use the ImageNet images from the CINIC dataset \cite{darlow2018cinic}. 
CINIC contains 2100 images randomly selected for each of the CIFAR-10 categories from the ImageNet dataset.
Hence, the performance of models trained on CIFAR-10 on these 21000 images can be considered as an approximation for a model's out-of-distribution generalization performance.

\subsection{Adversarial Robustness}
Deep Neural Networks are highly vulnerable to carefully crafted imperceptible perturbations designed to fool a neural network by an adversary \cite{biggio2013evasion,szegedy2013intriguing}.
This vulnerability poses a real threat to the deep learning model's deployment in the real world \cite{kurakin2016adversarial}.
Robustness to these adversarial attacks has therefore gained a lot of traction in the research community and progress has been made to better evaluate robustness to adversarial attacks \cite{carlini2017towards,goodfellow2014explaining,moosavi2016deepfool} and defend the models against these attacks \cite{madry2017towards,zhang2019theoretically}.

To evaluate the adversarial robustness of models in this study, we use the Projected Gradient Descent (PGD) attack from Madry \etal \cite{madry2017towards}.
The PGD attack initializes the adversarial image with the original image with the addition of a random noise within some epsilon bound, $\epsilon$. For each step, it takes the loss with respect to the input image and moves in the direction of loss with the step size and then clips it within the epsilon bound and the range of the valid image. 
In all of our experiments, we use $l_\infty$ attack with 0.031 epsilon bound, 0.03 step size. We use the notation PGD-N for an N steps PGD attack and report the worst performance of 5 random initialization runs.

\subsection{Natural Robustness}
While robustness to adversarial attack is important from a security perspective, it is an instance of a worst-case distribution shift.
The model also needs to be robust to naturally occurring perturbations which it will encounter frequently in the test environment.
Recent works have shown that Deep Neural Networks are also vulnerable to commonly occurring perturbations in the real world which are far from the adversarial examples manifold.
Hendrycks~\etal \cite{hendrycks2019natural} curated a set of real-world, unmodified, and naturally occurring examples that cause classifier accuracy to degrade sharply.
Gu \etal~\cite{gu2019using} measured model's robustness to the minute transformations found across video frames which they refer to as natural robustness and found state-of-the-art classifiers to be brittle to these transformations. 
In our study we use robustness to the common corruptions and perturbations proposed by Hendrycks \etal~\cite{hendrycks2019benchmarking} in CIFAR-C as a proxy for natural robustness.
We evaluate average robustness to the 19 distortions across 5 severity levels. Furthermore, we calculate mean Corruption Accuracy (mCA) over the 19 distortions.
Following \cite{gu2019using} we use a fine-grained measure of natural robustness, by computing accuracy on the corrupted image conditional on the clean image being classified correctly.

\section{Experimental Setup}
To study the effect of injecting noise in the knowledge distillation framework, we use Hinton method \cite{hinton2015distilling} which trains the student by minimizing the Kullback–Leibler divergence between the smoother output probabilities of the student and teacher. 
In all of our experiments we use the balancing parameter $\alpha=0.9$ and softmax temperature $\tau=4$ which are commonly used in knowledge distillation literature \cite{tung2019similarity,zagoruyko2016paying}.
We conduct our experiments on Wide Residual Networks (WRN) \cite{zagoruyko2016wide}. 
Unless otherwise stated, we normalize the images between 0 and 1 and use the standard training scheme as used in \cite{tung2019similarity,zagoruyko2016paying}: SGD with 0.9 momentum; 200 epochs; batch size 128; and an initial learning rate of 0.1, decayed by a factor of 0.2 at epochs 60, 120 and 150.
We conduct our experiments on CIFAR-10 \cite{krizhevsky2009learning} and SVHN \cite{netzer2011reading}, with WRN-40-2 with 2.2M parameters as the teacher, and WRN-16-2 with 0.7M parameters as the student because of their pervasiveness in literature. We also compare all the methods with the baseline which refers to WRN-16-2 trained alone with standard cross-entropy loss.
In all of our experiments, we train each model for five different seed values.
For the teacher, we select the model with the highest test accuracy and then use it to train the student again for five different seed values and report the mean and 1 std for our evaluation metrics.
For a fair comparison, we train the knowledge distillation methods under the same experimental setup using the publicly available code.

\section{Empirical study of Noises}
In this section, we propose injecting different types of noise in the student-teacher collaborative learning framework and analyze their effect on the performance of the student.

\subsection{Fickle Teacher}
Trial-to-trial response variability in the brain can be considerably different across stimuli, suggesting that it could also provide an important contribution to the information conveyed by the neural responses about the stimuli \cite{scaglione2011trial}.
Similarly, in deep neural networks, dropout \cite{srivastava2014dropout} which randomly switches off a subgroup of hidden units results in response variability and can be used to obtain principled uncertainty estimates for an input image \cite{gal2015dropout}.
We, therefore, propose to use the response variability resulting from keeping the dropout in the teacher model active to simulate the trial-to-trial response variability in the brain.
Fickle Teacher (FT) involves first training the teacher with dropout and in the subsequent step keeping the dropout active in the teacher while distilling knowledge to the student.
This results in variability in the supervision signal from the teacher to the student for the same input across different epochs, thereby exposing the student to its uncertainty.
It is important to note that the student itself does not use dropout.
The teacher, on the other hand, not only uses dropout during its training but also keeps it active when providing supervision to the student in order to provide additional information about the uncertainty of its prediction on a particular data point.

\begin{table*}[tbhp]
\caption{Fickle Teacher (FT) consistently improves both in-distribution and out-of-distribution (CINIC) generalization on CIFAR-10 as well as robustness to common corruptions (mCA). Similar generalization gain is observed on SVHN. The dropout rate, x, used for training the teacher is indicated by FT-x. The best performing models are in bold.}
\label{tab:fickle_teacher}
\centering
\begin{tabular}{|l|cccc||cc|}
\hline
& \multicolumn{4}{c||}{\bf CIFAR-10} & \multicolumn{2}{c|}{\bf SVHN} \\ \cline{2-7}
\bf Method                          & Teacher &   Test Acc.        &    CNIC Acc.       &          mCA       & Teacher &    Test Acc.       \\ \hline \hline
\bf Baseline                         & -       & 93.95$\pm$0.18     & 68.89$\pm$0.08     & 74.38$\pm$0.67     &     -   & 96.14$\pm$0.15     \\ \hline
\bf Hinton                          & 95.11   & 94.28$\pm$0.09     & 69.13$\pm$0.29     & 74.57$\pm$0.29     & 96.78   & 96.80$\pm$0.08     \\
\bf AT\cite{zagoruyko2016paying}    & 95.11   & 94.50$\pm$0.18     & 69.23$\pm$0.18     & 74.70$\pm$0.58     & 96.78   & 96.28$\pm$0.13     \\
\bf SP\cite{tung2019similarity}     & 95.11   & 94.64$\pm$0.17     & 69.39$\pm$0.32     & 74.93$\pm$0.43     & 96.78   & 96.61$\pm$0.06     \\
\bf RKD-D\cite{park2019relational}  & 95.11   & 94.42$\pm$0.15     & 69.34$\pm$0.17     & 74.75$\pm$0.60     & 96.78   & 96.49$\pm$0.05     \\
\bf RKD-A\cite{park2019relational}  & 95.11   & 94.62$\pm$0.14     & 69.57$\pm$0.17     & \bf 75.33$\pm$0.22     & 96.78   & 96.57$\pm$0.06     \\
\bf RKD-DA\cite{park2019relational} & 95.11   & 94.52$\pm$0.11     & 69.43$\pm$0.23     & 74.93$\pm$0.43     & 96.78     &   96.58$\pm$0.03   \\  \hline
\bf FT-0.1                          & 95.19   & 94.43$\pm$0.15     & 69.49$\pm$0.23     & 74.99$\pm$0.48     & 96.90   & 96.79$\pm$0.03     \\
\bf FT-0.2                          & \textbf{95.38}   & 94.46$\pm$0.16     & 69.59$\pm$0.13     & 74.61$\pm$0.41     & 96.85   & 96.74$\pm$0.06     \\
\bf FT-0.3                          & 95.12   & 94.56$\pm$0.14     & 69.84$\pm$0.22     & 75.06$\pm$0.14     & 96.94   & 96.90$\pm$0.07     \\
\bf FT-0.4                          & 95.18   & \bf 94.67$\pm$0.09 & 69.50$\pm$0.21     &  75.09$\pm$0.51 & 96.95   & 96.93$\pm$0.07     \\
\bf FT-0.5                          & 94.88   & 94.50$\pm$0.23     & \bf 69.95$\pm$0.25 & 74.67$\pm$0.30     & \textbf{97.00}   & \bf 97.09$\pm$0.02 \\ \hline 
\end{tabular}
\end{table*}

\begin{table*}[tb]
\caption{Fickle Teacher (+FT) better complements the other distillation methods compared to Hinton (+H). We train the models on CIFAR-10 using the same experimental setup as for the original methods. We use FT-0.4 for all the +FT experiments. The best performing models are in bold.}
\label{tab:FT_vs_Hinton}
\centering
\begin{tabular}{|l|ccc||ccc|}
\hline
 & \multicolumn{3}{c||}{Test Acc.} & \multicolumn{3}{c|}{CNIC Acc.} \\\cline{2-7}
\bf Method                            & Original            & +H             & +FT            & Original            & +H              & +FT\\ \hline
{\bf AT}\cite{zagoruyko2016paying}    & 94.50$\pm$0.18 & 94.63$\pm$0.24 & \bf 94.83$\pm$0.05 & 69.23$\pm$0.18 & \bf 69.57$\pm$0.13 & 69.46$\pm$0.23 \\
{\bf SP}\cite{tung2019similarity}     & 94.64$\pm$0.17 & 94.39$\pm$0.18 & \bf 94.66$\pm$0.09 & \bf 69.39$\pm$0.32 & 69.33$\pm$0.09 & 69.36$\pm$0.31 \\
{\bf RKD-D}\cite{park2019relational}  & 94.42$\pm$0.15 & 94.39$\pm$0.15 & \bf 94.76$\pm$0.04 & 69.34$\pm$0.17 & 69.21$\pm$0.11 & \bf 69.60$\pm$0.07 \\
{\bf RKD-A}\cite{park2019relational}  & 94.62$\pm$0.14 & 94.38$\pm$0.21 & \bf 94.68$\pm$0.06 & 69.57$\pm$0.17 & 69.15$\pm$0.17 & \bf 69.68$\pm$0.33 \\
{\bf RKD-DA}\cite{park2019relational} & 94.52$\pm$0.11 & 94.43$\pm$0.04 & \bf 94.59$\pm$0.14 & 69.43$\pm$0.23 & 69.36$\pm$0.05 & \bf 69.62$\pm$0.17 \\ \hline
\end{tabular}
\end{table*}

We systematically change the dropout rate used for training the teacher and study its effect on the generalization performance of the student.
Because of the variability in the teacher's supervision signal, the student needs to be trained for more epochs in order for it to converge and be effectively exposed to the uncertainty of the teacher.
We use the same initial learning rate of 0.1 and a decay factor of 0.2 as per the standard training scheme.
For a dropout rate of 0.1 and 0.2, we train for 250 epochs and reduce the learning rate at 75, 150, and 200 epochs.
For dropout rate 0.3, we train for 300 epochs and reduce the learning rate at 90, 180, and 240 epochs.
Finally for a dropout rate of 0.4 and 0.5, due to the increased variability, we train for 350 epochs and reduce the learning rate at 105, 210, and 280 epochs.
We show the efficacy of our approach by comparing it with the state-of-the-art knowledge distillation methods under the same experimental settings. Following the parameters used in the paper, we use $\beta=1000$ for Attention (AT) \cite{zagoruyko2016paying}, $\gamma=3000$ for Similarity Preserving (SP) \cite{tung2019similarity} whereas for Relational Knowledge Distillation (RKD), we use $\lambda_{RKD-D}=25$ and $\lambda_{RKD-A}=50$.

Table \ref{tab:fickle_teacher} shows that FT improves the in-distribution and out-of-distribution generalization on CIFAR-10 as well as the robustness to common corruptions.
Notably, even when the accuracy of the teacher decreases after a dropout rate of 0.2, the student accuracy still improves up to a dropout rate of 0.4. FT provides similar generalization gains for SVHN. For FT-0.5, the student even outperforms the teacher.

Furthermore, since it is a common practice to add Hinton loss on top of other distillation methods, Table \ref{tab:FT_vs_Hinton} compares the effect of adding Hinton vs FT on top of the other distillation methods.
The higher generalization gains with FT across all the distillation methods show that it better complements these methods compared to Hinton.
For relational knowledge distillation methods (SP, all variants of RKD) where adding Hinton loss reduces performance, FT provides further improvement over the original method.

FT maintains the simplicity and versatility of the Hinton method and provides even higher generalization gains over the recently proposed knowledge distillation variants which not only adds more constraints to student training, hence limiting their versatility but also require a lot of parameter tuning. Also, FT can be easily added on top of other distillation methods to further improve the performance of the model. The effectiveness of FT in improving the generalization of the model motivates further exploration into techniques that add noise so that it encodes the uncertainty in the supervision signal.

\subsection{Soft Randomization}
Pinot \etal \cite{pinot2019theoretical} show that the injection of noise drawn from an exponential family such as Gaussian or Laplace noise leads to guaranteed robustness to adversarial attack. 
However, this improved adversarial robustness comes at the cost of significant loss in generalization.
Some studies even argue that there is an inherent trade-off between robustness and generalization and consider them as contradictory goals \cite{ilyas2019adversarial,tsipras2018robustness}. 
Therefore, an important consideration for methods proposed to increase the adversarial robustness is to reduce this loss in generalization.

Since knowledge distillation provides an opportunity to combine multiple sources of information, we hypothesize that combining information from a teacher with high generalization while training the student to be robust to noisy input can reduce the trade-off.
The extra supervision signal from the teacher acts as a regularizer which encourages the student to align its feature distribution with the teacher.
This effectively adds a prior, encouraging the student to learn semantically relevant features which are robust to the spurious directions introduced by Gaussian noise.
It can also be considered as distilling knowledge from the clean domain to the noisy domain.

\begin{table*}[tb]
\caption{Soft Randomization (SR) consistently achieves higher in-distribution and out-of-distribution generalization on CIFAR-10 and for the majority of the noise intensities on SVHN compared to Gaussian Augmentation (GA). The best performing models are in bold.}
\label{tab:sr_gen}
\centering
\begin{tabular}{|c|cc|cc||cc|}
\hline
 & \multicolumn{2}{c|}{CIFAR-10 Test Acc.} & \multicolumn{2}{c||}{CINIC Acc.} & \multicolumn{2}{c|}{SVHN Test Acc.} \\ \cline{2-7}
Sigma & GA & SR & GA & SR & GA & SR \\ \hline
0.01 & 93.80$\pm$0.25 & \bf 94.29$\pm$0.13 & 69.08$\pm$0.21 & \bf 69.19$\pm$0.21 & 96.14$\pm$0.11 & \bf 96.78$\pm$0.10\\
0.02 & 93.65$\pm$0.10 & \bf 94.07$\pm$0.16 & 68.98$\pm$0.14 & \bf 69.11$\pm$0.40 & 96.29$\pm$0.10 & \bf 96.79$\pm$0.12 \\
0.03 & 93.14$\pm$0.20 & \bf 93.53$\pm$0.29 & 68.59$\pm$0.22 & \bf 68.77$\pm$0.26 & 96.24$\pm$0.08 & \bf 96.77$\pm$0.05 \\
0.04 & 92.67$\pm$0.10 & \bf 93.04$\pm$0.18 & 68.05$\pm$0.19 & \bf 68.33$\pm$0.37 & 96.08$\pm$0.11 & \bf 96.72$\pm$0.09 \\
0.05 & 92.14$\pm$0.26 & \bf 92.57$\pm$0.22 & 67.52$\pm$0.16 & \bf 68.07$\pm$0.27 & 95.93$\pm$0.16 & \bf 96.56$\pm$0.08 \\ \hline
0.1  & 89.09$\pm$0.26 & \bf 89.88$\pm$0.14 & 64.89$\pm$0.20 & \bf 65.44$\pm$0.37 & 95.55$\pm$0.22 & \bf 96.01$\pm$0.14 \\
0.2  & 83.41$\pm$0.21 & \bf 84.07$\pm$0.30 & 59.31$\pm$0.20 & \bf 60.08$\pm$0.17 & 93.22$\pm$0.20 & \bf 93.39$\pm$0.22 \\
0.3  & 78.00$\pm$0.49 & \bf 78.51$\pm$0.48 & 54.55$\pm$0.34 & \bf 55.19$\pm$0.25 & \bf 89.75$\pm$0.29 & 89.70$\pm$0.36 \\
0.4  & 72.88$\pm$0.46 & \bf 73.35$\pm$0.37 & 50.30$\pm$0.18 & \bf 51.03$\pm$0.36 & \bf 85.51$\pm$0.35 & 85.08$\pm$0.31 \\
0.5  & 68.39$\pm$0.45 & \bf 68.95$\pm$0.21 & 46.75$\pm$0.31 & \bf 47.44$\pm$0.13 & \bf 80.48$\pm$0.53 & 79.69$\pm$0.83 \\ \hline
\end{tabular}
\end{table*}

\begin{figure}[tb]
 \centering
 \includegraphics[trim=2cm 15cm 15cm .5cm clip, width=.8\columnwidth]{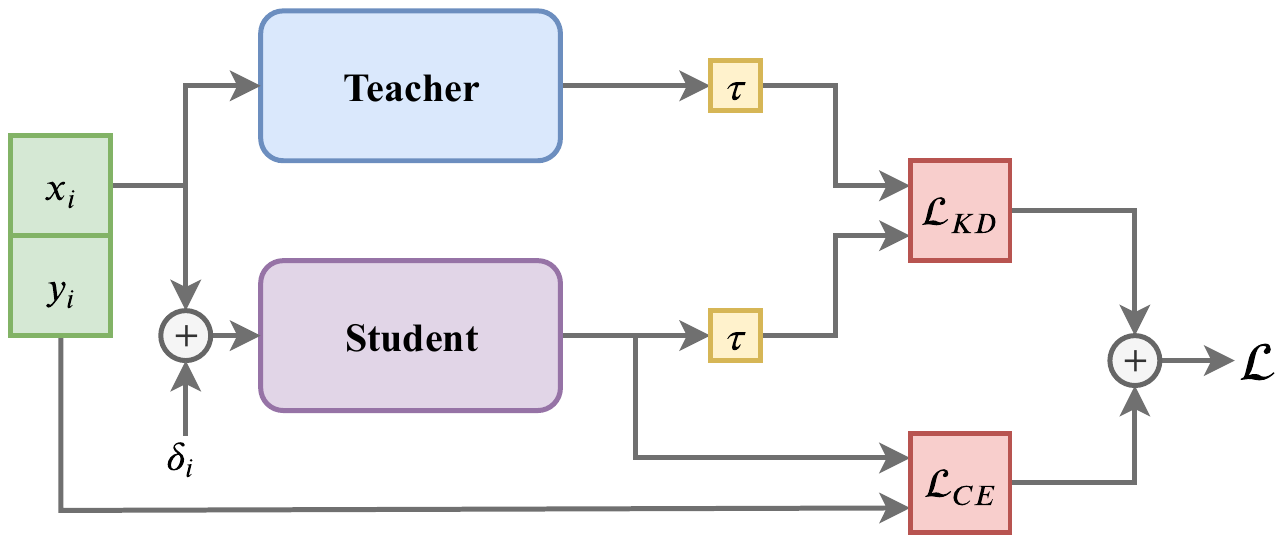}
 \caption{Soft Randomization uses supervision from a static teacher (trained on clean data) to train the student model effectively using Gaussian data augmentation. For training the student, SR minimizes the KL divergence between student's response on noisy input and teacher's response on the clean input in addition to minimizing the cross-entropy loss on the noisy input.}
 \label{fig:SR}
\end{figure}

To test the hypothesis, we propose a novel technique for improving robustness to input variability in the student which utilizes the teacher trained on clean data, to train the student on noisy data.
Here, we minimize the dissimilarity between the student’s distribution on noisy data with the teacher’s distribution on clean data (Figure~\ref{fig:SR}).
Therefore, loss function for SR adapts the vanilla knowledge distillation loss (Eq. \ref{equ_KD}):
\begin{equation}\label{equ_SR}
 \mathcal{L}=(1-\alpha)\mathcal{L}_{CE}(S(x+\delta),y)+\alpha\tau^2D_{KL}(S^\tau(x+\delta)||T^\tau(x))
\end{equation}
where $\delta\sim\mathcal{N}(0,\sigma^2)$ is white Gaussian noise. 

\begin{figure}[tb]
\centering
\includegraphics[width=1\columnwidth]{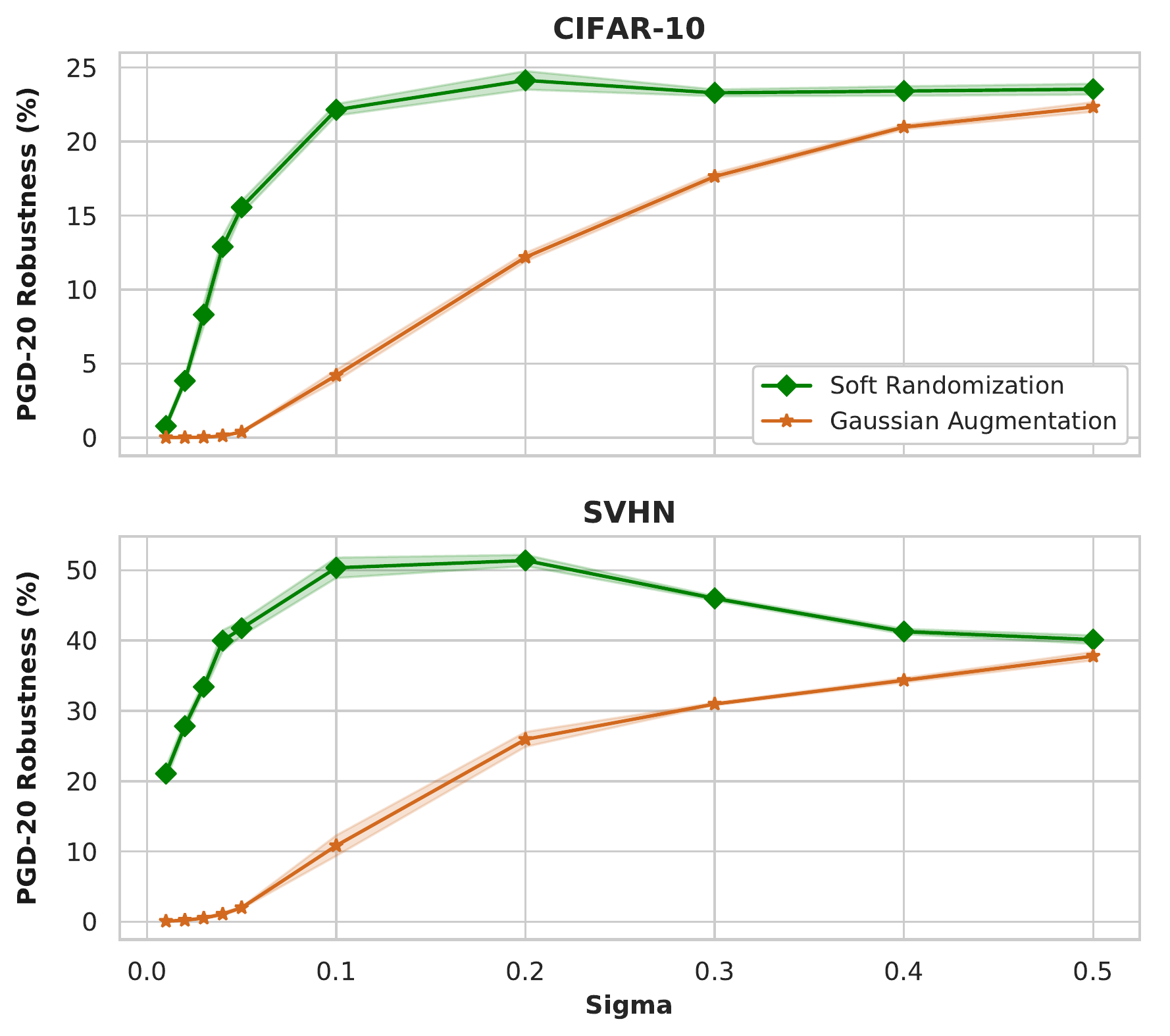}
\caption{Soft Randomization consistently achieves higher adversarial robustness compared to Gaussian Augmentation. For lower noise intensities, Soft Randomization provides significantly higher robustness to PGD-20 attacks. We observe peak robustness at 0.2 where the model achieves both high generalization and robustness. Shaded regions show 1 std. For values, see Supplementary Material Table S1.}
\label{fig:sr_robustness}
\end{figure}

We train Soft Randomization (SR) for both low noise and high noise intensities.
We compare Soft Randomization to the compact model (WRN-16-2) trained alone with Gaussian data augmentation, referred to as Gaussian augmentation (GA).
Figure~\ref{fig:sr_robustness} shows that SR consistently improves the adversarial robustness of the student over GA for both datasets. Especially for lower noise intensities, SR outperforms GA by a considerable margin, indicating that SR enables training a robust model even with low noise intensities which have the advantage of higher generalization performance as well. Table \ref{tab:sr_gen} shows the corresponding generalization performance of the models. To complement the robustness gains, SR consistently achieves better in-distribution and out-of-distribution generalization for all $\sigma$ values compared to GA on CIFAR-10 and the majority of noise intensities on SVHN. For $\sigma=0.05$, SR achieves 15.56\% robustness to PGD-20 attack and 92.57\% test accuracy compared to only 0.38\% robustness and 92.14\% test accuracy for GA on CIFAR-10. For SVHN, the difference is even more pronounced, with SR achieving 93.39\% generalization with 51.39\% robustness compared to GA's 93.22\% generalization and 25.94\% robustness for $\sigma=0.2$.

\begin{table*}[tb]
\caption{Comparison of Soft Randomization (SR) and Gaussian Augmentation (GA) on PGD attacks with fixed epsilon bound ($\epsilon=8/255$) and an increasing number of iterations. SR consistently provides higher robustness against the PGD attacks of varying strengths. As expected, models trained with higher noise intensity, defend better against stronger attacks. The best performing models are in bold.}
\label{tab:SR_gradient_obf_pgd}
\centering
\resizebox{\textwidth}{!}{
\begin{tabular}{|l|l|l|cccccccc|}
\hline
 & $\sigma$ & & 1 & 5 & 10 & 15 & 20 & 100 & 200 & 1000\\ \hline
\multirow{8}{*}{\rotatebox[origin=c]{90}{CIFAR-10}} 
& 0.02 & GA & 72.85$\pm$0.19 & 5.08$\pm$0.22 & 0.10$\pm$0.02 & 0.00$\pm$0.01 & 0.00$\pm$0.00 & 0.00$\pm$0.00 & 0.00$\pm$0.00 & 0.00$\pm$0.00 \\
& & SR & \textbf{78.15$\pm$0.17} & \textbf{39.06$\pm$0.63} & \textbf{18.02$\pm$0.85} & \textbf{8.07$\pm$0.37} & \textbf{3.83$\pm$0.28} & \textbf{0.11$\pm$0.05} & \textbf{0.06$\pm$0.04} & \textbf{0.03$\pm$0.03} \\ \cline{2-11}
& 0.05 & GA & 81.24$\pm$0.23 & 26.53$\pm$0.71 & 4.91$\pm$0.38 & 1.10$\pm$0.10 & 0.38$\pm$0.04 & 0.08$\pm$0.02 & 0.07$\pm$0.02 & 0.07$\pm$0.01 \\
& & SR & \textbf{83.20$\pm$0.19} & \textbf{49.74$\pm$0.34} & \textbf{31.30$\pm$0.51} & \textbf{21.18$\pm$0.39} & \textbf{15.56$\pm$0.44} & \textbf{3.80$\pm$0.69} & \textbf{2.80$\pm$0.62} & \textbf{1.91$\pm$0.53} \\ \cline{2-11}
& 0.2 & GA & 76.62$\pm$0.21 & 49.12$\pm$0.32 & 26.62$\pm$0.24 & 16.06$\pm$0.33 & 12.19$\pm$0.29 & 9.83$\pm$0.22 & 9.73$\pm$0.23 & 9.68$\pm$0.23 \\
& & SR & \textbf{78.05$\pm$0.22} & \textbf{53.64$\pm$0.42} & \textbf{35.92$\pm$0.48} & \textbf{28.05$\pm$0.65} & \textbf{24.14$\pm$0.63} & \textbf{16.87$\pm$0.57} & \textbf{16.36$\pm$0.58} & \textbf{15.98$\pm$0.52} \\ \cline{2-11}
& 0.5 & GA & 63.81$\pm$0.65 & 47.51$\pm$0.35 & 33.35$\pm$0.34 & 25.51$\pm$0.38 & 22.34$\pm$0.33 & 20.87$\pm$0.35 & 20.80$\pm$0.37 & 20.79$\pm$0.37 \\
& & SR & \textbf{64.53$\pm$0.32} & \textbf{48.11$\pm$0.27} & \textbf{34.16$\pm$0.20} & \textbf{26.52$\pm$0.38} & \textbf{23.54$\pm$0.37} & \textbf{21.85$\pm$0.29} & \textbf{21.81$\pm$0.27} & \textbf{21.79$\pm$0.27} \\ \hline
 
\multirow{8}{*}{\rotatebox[origin=c]{90}{SVHN}} 
& 0.02 & GA & 86.77$\pm$0.43 & 24.24$\pm$1.51 & 2.97$\pm$0.47 & 0.59$\pm$0.14 & 0.21$\pm$0.06 & 0.03$\pm$0.01 & 0.02$\pm$0.01 & 0.02$\pm$0.01 \\
& & SR & \textbf{90.92$\pm$0.32} & \textbf{67.50$\pm$0.66} & \textbf{49.11$\pm$0.79} & \textbf{35.87$\pm$0.83} & \textbf{27.82$\pm$0.83} & \textbf{6.20$\pm$0.44} & \textbf{4.32$\pm$0.36} & \textbf{2.71$\pm$0.30} \\ \cline{2-11}
& 0.05 & GA & 90.01$\pm$0.29 & 45.14$\pm$0.92 & 12.48$\pm$0.73 & 4.05$\pm$0.29 & 1.99$\pm$0.13 & 0.51$\pm$0.07 & 0.43$\pm$0.06 & 0.37$\pm$0.05 \\
& & SR & \textbf{92.38$\pm$0.21} & \textbf{72.74$\pm$0.60} & \textbf{58.21$\pm$0.69} & \textbf{47.97$\pm$0.91} & \textbf{41.75$\pm$1.06} & \textbf{17.72$\pm$0.97} & \textbf{14.76$\pm$0.90} & \textbf{11.68$\pm$0.88} \\ \cline{2-11}
& 0.2 & GA & 89.56$\pm$0.33 & 67.44$\pm$0.59 & 43.24$\pm$0.75 & 30.08$\pm$0.88 & 25.94$\pm$1.05 & 17.94$\pm$1.02 & 17.57$\pm$1.04 & 17.40$\pm$1.02 \\
& & SR & \textbf{89.80$\pm$0.24} & \textbf{71.85$\pm$0.43} & \textbf{59.11$\pm$0.67} & \textbf{52.23$\pm$0.88} & \textbf{51.39$\pm$0.80} & \textbf{36.51$\pm$0.74} & \textbf{35.33$\pm$0.70} & \textbf{34.48$\pm$0.69} \\ \cline{2-11}
& 0.5 & GA & \textbf{76.41$\pm$0.54} & \textbf{60.48$\pm$0.50} & \textbf{44.43$\pm$0.48} & 34.70$\pm$0.59 & 37.78$\pm$0.60 & 27.02$\pm$0.50 & 26.85$\pm$0.49 & 26.82$\pm$0.48 \\
& & SR & 75.50$\pm$0.81 & 59.24$\pm$0.53 & 44.10$\pm$0.23 & \textbf{35.74$\pm$0.22} & \textbf{40.13$\pm$0.61} & \textbf{28.07$\pm$0.32} & \textbf{27.96$\pm$0.33} & \textbf{27.94$\pm$0.33} \\ \hline

\end{tabular}}
\end{table*}

\begin{table*}[tb]
\caption{Comparison of SR and GA on PGD attacks with a fixed number of steps (20) and increasing epsilon bounds. The robustness effectively goes to zero as the epsilon budget increases which shows that gradient-based attacks perform as expected on SR. SR consistently provides higher robustness against the PGD attacks of increasing strengths. The best performing models are in bold.}
\label{tab:SR_gradient_obf_eps}
\centering
\resizebox{\textwidth}{!}{
\begin{tabular}{|l|l|l|cccccccc|}
\hline
 & $\sigma$ &  & 1 & 2 & 10 & 20 & 25 & 50 & 100 & 200 \\ \hline
 \multirow{8}{*}{\rotatebox[origin=c]{90}{CIFAR-10}} 
& 0.02 & GA & 60.27$\pm$0.59 & 19.60$\pm$0.54 & 0.00$\pm$0.00 & 0.00$\pm$0.00 & 0.00$\pm$0.00 & 0.00$\pm$0.00 & 0.00$\pm$0.00 & 0.00$\pm$0.00 \\
& & SR & \textbf{66.35$\pm$0.37} & \textbf{35.85$\pm$0.63} & \textbf{2.42$\pm$0.18} & \textbf{0.40$\pm$0.05} & \textbf{0.19$\pm$0.02} & \textbf{0.02$\pm$0.01} & 0.00$\pm$0.00 & 0.00$\pm$0.00 \\ \cline{2-11}
& 0.05 & GA & 74.51$\pm$0.35 & 48.26$\pm$0.71 & 0.14$\pm$0.02 & 0.00$\pm$0.00 & 0.00$\pm$0.00 & 0.00$\pm$0.00 & 0.00$\pm$0.00 & 0.00$\pm$0.00 \\
& & SR & \textbf{76.94$\pm$0.21} & \textbf{55.01$\pm$0.39} & \textbf{13.03$\pm$0.47} & \textbf{7.37$\pm$0.60} & \textbf{5.32$\pm$0.54} & \textbf{0.38$\pm$0.09} & \textbf{0.02$\pm$0.04} & 0.00$\pm$0.00 \\ \cline{2-11}
& 0.2 & GA & 73.74$\pm$0.18 & 62.56$\pm$0.13 & 7.57$\pm$0.18 & 3.09$\pm$0.11 & 2.53$\pm$0.11 & 1.23$\pm$0.12 & 0.11$\pm$0.03 & 0.00$\pm$0.00 \\
& & SR & \textbf{75.47$\pm$0.16} & \textbf{64.91$\pm$0.23} & \textbf{20.98$\pm$0.58} & \textbf{16.91$\pm$0.75} & \textbf{16.12$\pm$0.85} & \textbf{13.57$\pm$0.81} & \textbf{4.90$\pm$0.58} & \textbf{0.03$\pm$0.03} \\ \cline{2-11}
& 0.5 & GA & 62.14$\pm$0.65 & 55.73$\pm$0.50 & 16.99$\pm$0.37 & 9.32$\pm$0.34 & 8.17$\pm$0.27 & 6.03$\pm$0.27 & 3.84$\pm$0.20 & 0.55$\pm$0.05 \\
& & SR & \textbf{62.97$\pm$0.42} & \textbf{56.43$\pm$0.33} & \textbf{18.56$\pm$0.35} & \textbf{12.00$\pm$0.31} & \textbf{11.00$\pm$0.31} & \textbf{9.16$\pm$0.31} & \textbf{6.60$\pm$0.25} & \textbf{1.02$\pm$0.08} \\ \hline

\multirow{8}{*}{\rotatebox[origin=c]{90}{SVHN}} 
& 0.02 & GA & 77.80$\pm$0.73 & 40.52$\pm$1.66 & 0.08$\pm$0.03 & 0.01$\pm$0.01 & 0.01$\pm$0.00 & 0.00$\pm$0.00 & 0.00$\pm$0.00 & 0.00$\pm$0.00 \\
& & SR & \textbf{85.82$\pm$0.35} & \textbf{67.78$\pm$0.48} & \textbf{23.14$\pm$0.83} & \textbf{14.53$\pm$0.60} & \textbf{11.59$\pm$0.49} & \textbf{2.98$\pm$0.19} & \textbf{0.18$\pm$0.04} & 0.00$\pm$0.00 \\ \cline{2-11}
& 0.05 & GA & 85.59$\pm$0.40 & 63.31$\pm$0.64 & 0.88$\pm$0.11 & 0.19$\pm$0.04 & 0.12$\pm$0.03 & 0.02$\pm$0.01 & 0.00$\pm$0.00 & 0.00$\pm$0.00 \\
& & SR & \textbf{89.30$\pm$0.25} & \textbf{76.25$\pm$0.48} & \textbf{36.56$\pm$1.07} & \textbf{28.33$\pm$1.04} & \textbf{25.30$\pm$1.02} & \textbf{10.30$\pm$0.99} & \textbf{0.83$\pm$0.19} & 0.00$\pm$0.00 \\ \cline{2-11}
& 0.2 & GA & 87.71$\pm$0.32 & 79.01$\pm$0.53 & 17.47$\pm$1.01 & 9.11$\pm$0.91 & 7.87$\pm$0.86 & 5.11$\pm$0.74 & 1.96$\pm$0.35 & 0.03$\pm$0.01 \\
& & SR & \textbf{88.09$\pm$0.23} & \textbf{80.18$\pm$0.32} & \textbf{45.11$\pm$0.88} & \textbf{40.15$\pm$0.86} & \textbf{38.97$\pm$0.81} & \textbf{34.69$\pm$0.75} & \textbf{21.61$\pm$0.61} & \textbf{1.60$\pm$0.23} \\ \cline{2-11}
& 0.5 & GA & \textbf{74.86$\pm$0.56} & \textbf{68.63$\pm$0.50} & 23.41$\pm$0.50 & 13.80$\pm$0.38 & 12.32$\pm$0.28 & 9.38$\pm$0.20 & 6.57$\pm$0.16 & 1.49$\pm$0.05 \\
& & SR & 73.85$\pm$0.84 & 67.37$\pm$0.66 & \textbf{27.25$\pm$0.35} & \textbf{21.50$\pm$0.48} & \textbf{20.56$\pm$0.47} & \textbf{18.50$\pm$0.48} & \textbf{14.66$\pm$0.46} & \textbf{4.18$\pm$0.24} \\ \hline
\end{tabular}}
\end{table*}

To further analyze the robustness of SR, we evaluate the models on PGD attacks of varying strengths.
Table \ref{tab:SR_gradient_obf_pgd} shows the effect of increasing the number of iterations for a fixed epsilon bound ($\epsilon=8/255$). SR consistently provides higher robustness compared to GA across the PGD attacks of varying strengths and maintains its robustness level after 100 steps. To further show that gradient-based methods are indeed effective on the proposed method, Table \ref{tab:SR_gradient_obf_eps} shows that increasing the allowed perturbation budget (epsilon) for a fixed number of iterations (20) effectively reduces the robustness of the models to 0.
This follows the analysis suggested by Lamb \etal \cite{lamb2019interpolated} and shows that SR does not suffer from gradient obfuscation~\cite{athalye2018obfuscated}.

\begin{figure*}[t!b]
 \centering
 \includegraphics[width=.95\textwidth]{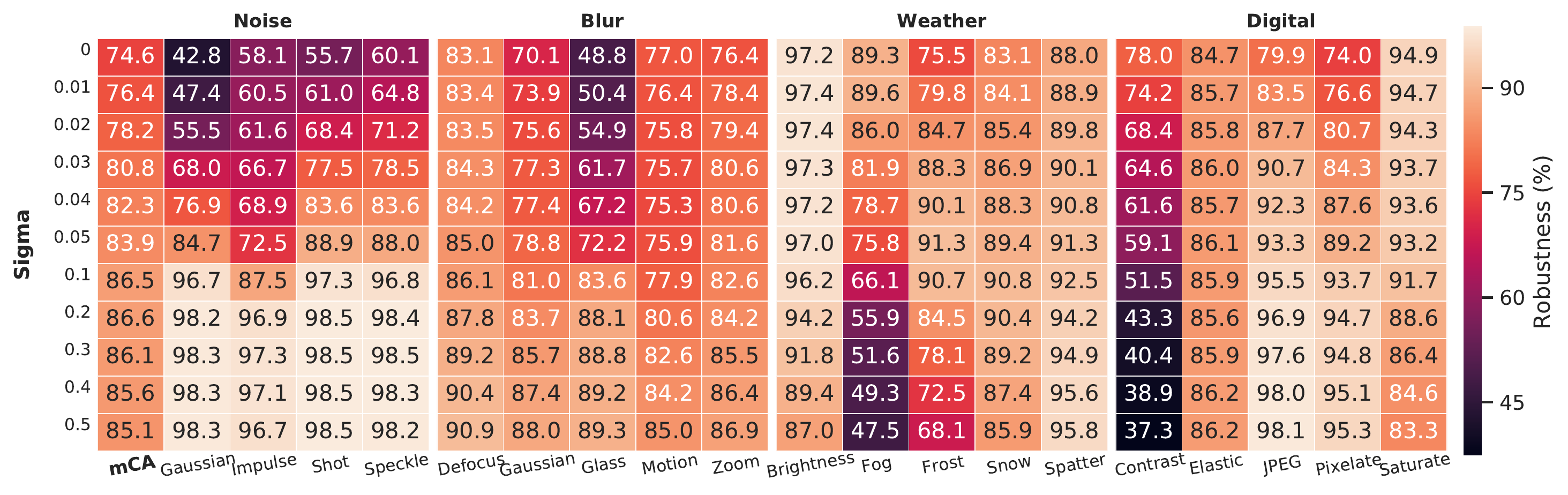}
 \caption{Soft Randomization consistently improves the average robustness to common corruptions (mCA). SR most notably improves the robustness to \textit{noise} and \textit{blurring distortions}.}
 \label{fig:gn_nat_rob}
\end{figure*}

Finally, Figure~\ref{fig:gn_nat_rob} shows the natural robustness of models trained on CIFAR-10 with SR. The mCA improves over the Hinton method for all noise intensities. While robustness drops most notably for color distortions (\textit{contrast}, \textit{brightness} and \textit{saturation}), robustness to \textit{noise} and \textit{blurring corruptions} improves as the Gaussian noise intensity increases.
We also observe changes in the effect of different noise intensities. For \textit{frost}, the robustness increases at a lower noise level and then decreases for higher intensities. 

The empirical results show that SR increases the capacity of the student to learn robust features even with lower noise intensities. This allows the use of lower noise intensity for increasing adversarial robustness while keeping the loss in generalization much lower compared to the GA model with a comparable robustness level. This essentially provides a better trade-off between robustness and generalization, enabling the training of compact students with both high robustness and generalization. SR, hence, provides greater flexibility in finding a suitable trade-off based on the application. The empirical results suggest that adding noise in the input data is more effective in a collaborative learning framework compared to a model in isolation.   

\begin{figure*}[tbh]
 \centering
   \includegraphics[width=0.9\textwidth]{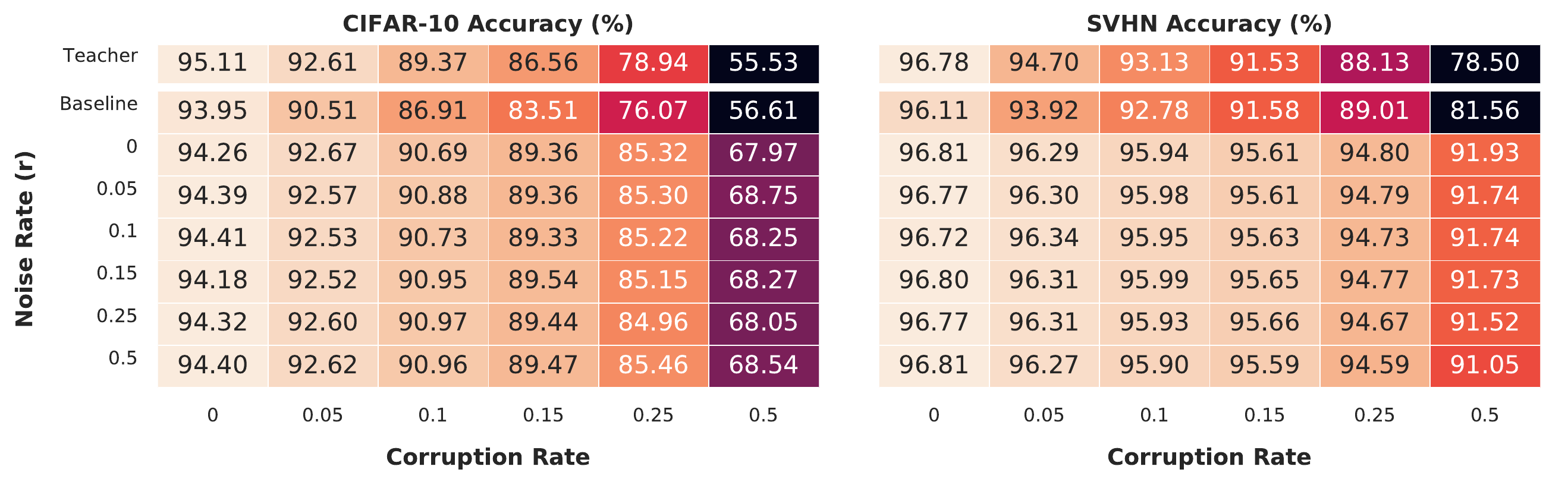}
 \caption{Generalization performance of Messy Collaboration (MC) with varying noise rates trained on corrupted labels. Knowledge distillation provides considerable generation gains compared to the Baseline and Teacher. MC, further, improves the generalization. For standard deviations, see Supplementary Material Tables S2 and S3.}
 \label{fig:mc_gen}
\end{figure*}

\subsection{Messy Collaboration}
Human decision-making shows systematic simplifications and deviations from the tenets of rationality (‘heuristics’) which may lead to sub-optimal decisional outcomes (‘cognitive biases’) \cite{korteling2018neural}. 
We believe this cognitive bias is manifested in deep neural networks in the form of memorization \cite{arpit2017closer} and over-generalization. This makes it even more challenging to learn efficiently on real-world datasets with some fraction of corrupted labels (mislabeled) which have been shown to adversely affect the model performance~\cite{frenay2013classification}. Furthermore, in order to utilize the vast amount of open-source data available, researchers have proposed methods to generate labels automatically using user tags and keywords. However, these methods lead to noisy labels. Considering the abundance of noisy labels, it is imperative to develop methods that can effectively learn from noisy labels.

We propose a counter-intuitive regularization technique based on target variability to mitigate cognitive bias and subsequently reduce memorization in DNNs. We term this technique as Messy Collaboration (MC). While distilling knowledge from the teacher, for each sample in the training batch, with rate $r$, we randomly change the true labels to random targets sampled uniformly from the number of the classes. The target variability is added independently for each batch. We hypothesize that the target variability\footnote{We use the term {\it target variability} to refer to the random label corruption which we are introducing intentionally for each batch during training. Whereas, {\it noisy labels} refers to the inherent corruption in the labels which comes from incorrect annotations.} discourages memorization in DNNs and prevents overconfident predictions. Also, the soft targets from the teacher provide additional information about the similarities between the different classes which can mitigate the adverse effect of incorrect annotations and together with target variability improve the efficiency of DNNs to learn with noisy labels.

\begin{figure}[tb]
 \centering
   \includegraphics[width=1\columnwidth]{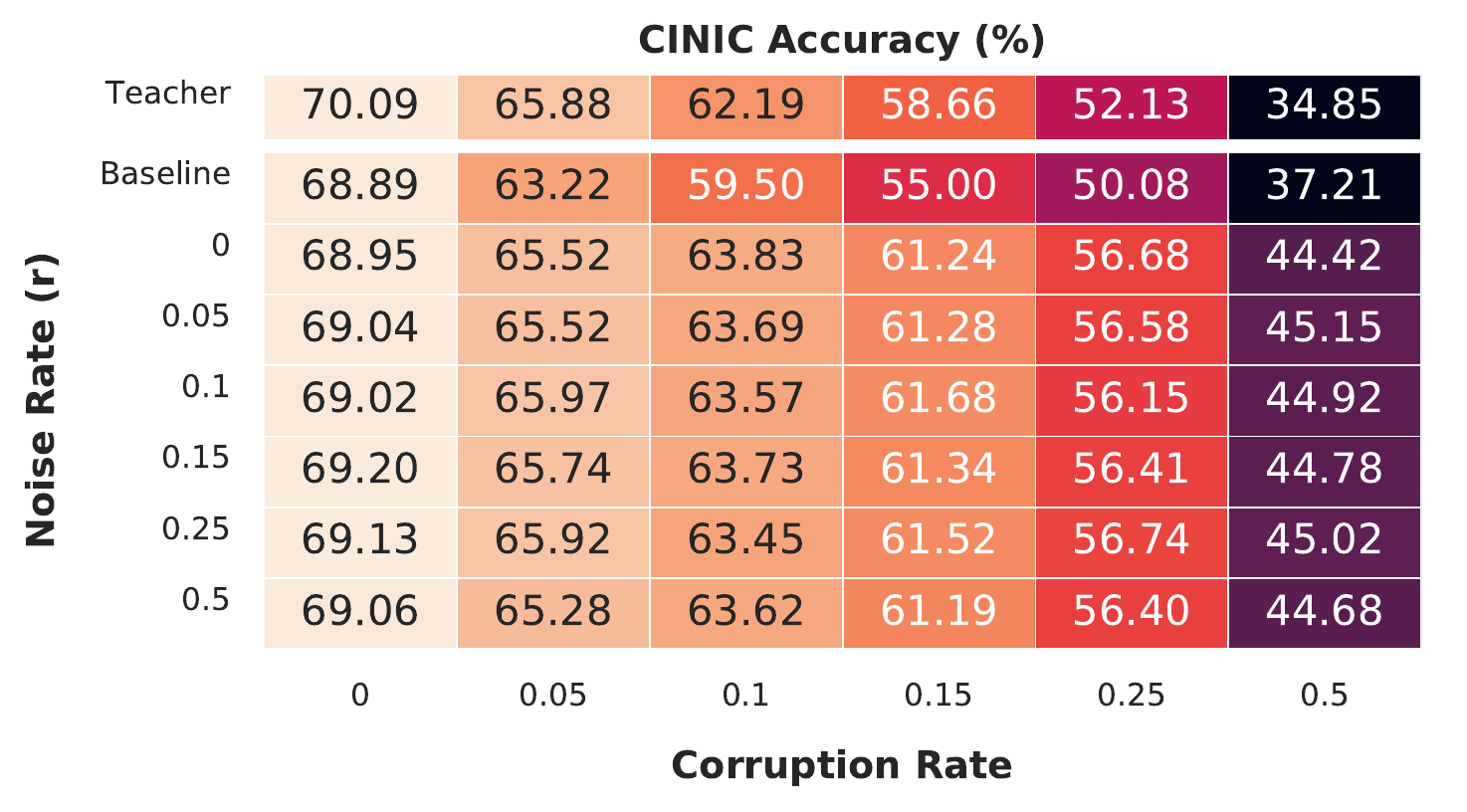}
 \caption{In addition to in-distribution generalization, Messy Collaboration also improves the out-of-distribution generalization over the Teacher and vanilla knowledge distillation under corrupted labels. For standard deviations, see Supplementary Material Table S4.}
 \label{fig:MC_gen_cinic}
\end{figure}

Here, we first show the effectiveness of knowledge distillation in learning with noisy labels at varying rates of label corruption on CIFAR-10 and SVHN and then further show that MC improves the generalization over vanilla knowledge distillation. Figure~\ref{fig:mc_gen} shows that the generalization drops as the corruption rate increases (cf. Teacher and Baseline). For all corruption levels, knowledge distillation (r=0) improves the generalization performance and even outperforms the teacher on both datasets.
The gain in generalization gets higher as the label corruption rate increases. 
It also shows the effect of varying the noise rate in MC for the different label corruption rates. On CIFAR-10, for label corruption rate 0.1 and higher, MC improves the generalization over the Baseline and teacher for all noise rates. 
For the majority of the corruption levels across the two datasets, MC further improves generalization over vanilla knowledge distillation. 
This shows that the target variability in MC makes the model more tolerant to label noise which allows efficient learning with noisy labels. The performance gain with MC over Hinton is less pronounced for SVHN, possibly because of the ease of the task. Figure \ref{fig:MC_gen_cinic} shows similar performance gains in out-of-distribution generalization as in-distribution for models trained on CIFAR-10.

The empirical results confirm that knowledge distillation is an effective framework for training models under noisy labels and provides consistent performance gain over models trained alone with cross-entropy. The further improvement with MC shows that target variability can discourage memorization in DNNs. This suggests that using noise in the target labels as a regularizer against memorization and over-confident predictions in DNNs is a promising direction.  

\section{Related Work}
A number of experimental and computational methods have reported the presence of noise in the nervous system and how it affects the function of the system \cite{faisal2008noise, scaglione2011trial}.
Analogously, noise has been shown to improve the training and performance of DNNs \cite{graves2011practical,li2017convergence,srivastava2014dropout,wan2013regularization,yim2017gift,zhou2017stochastic}. 
Furthermore, a family of randomization techniques that inject noise in the model both during training and inference time is proven to be effective to the adversarial attacks \cite{dhillon2018stochastic,liu2018towards,rakin2018parametric,xie2017mitigating}.
Randomized smoothing transforms any classifier into a new smooth classifier that has certifiable $l_2$-norm robustness guarantees \cite{cohen2019certified,lecuyer2018certified}. Label smoothing improves the performance of deep neural networks across a range of tasks \cite{pereyra2017regularizing, szegedy2016rethinking}.

Knowledge Distillation \cite{hinton2015distilling} has proven to be an effective framework for training compact models with the help of additional supervision signals from a larger static pre-trained model. A number of modifications have been proposed to the original formulation. AT~\cite{zagoruyko2016paying} proposed attention, defined as a set of spatial maps, as a mechanism for transferring knowledge to the student.  RKD~\cite{park2019relational} encourages the student to form the same relational structure in the output representation space with that of the teacher using two potential functions:  RKD-D measures the euclidean distance between two data samples while RKD-A measures the angle formed by the three data samples. RKD-DA is a combination of both losses. SP~\cite{tung2019similarity} preserves the pairwise similarities so that similar/dissimilar activations in the teacher produce similar/dissimilar activations in the student. While different ways of distilling knowledge to the student have been extensively studied \cite{sarfraz2020knowledge}, the role of noise in the knowledge distillation framework is not well studied.
Interestingly, Muller \etal \cite{muller2019does} report that label smoothing impairs knowledge distillation. 
On the contrary, we show that our biologically inspired technique of injecting noise into knowledge distillation, {\it Fickle Teacher}, consistently improves the generalization of the student. 
Furthermore, {\it Fickle Teacher} differs from the works of Bulo \etal \cite{bulo2016dropout} and Gurau \etal \cite{gurau2018dropout} in that instead of using the soft target distribution obtained by averaging Monte Carlo samples, we use the logits of the teacher model with dropout active directly as a source of uncertainty encoding noise for distilling knowledge to a compact student.

\section{Conclusion}
We demonstrated that noise can be used as a resource for learning in the knowledge distillation framework by injecting noise at multiple levels and extensively evaluating its effects on the performance of the model.
Inspired by trial-to-trial variability in the brain which can result from multiple noise sources, we introduced {\it Fickle Teacher} which exposes the student to its uncertainty using dropout. We show that the variability in the supervision signal improves both in-distribution and out-of-distribution generalization.
We further proposed {\it Soft Randomization} which utilizes input noise into the training of the student model. It involves matching the output distribution of the student on noisy data to the output distribution of the teacher on clean data. This considerably increases the capacity of the student to learn robust features and provides considerably higher adversarial robustness compared to the model trained alone with Gaussian data augmentation for lower noise intensities while also keeping the loss in generalization minimal.
Finally, {\it Messy Collaboration} employs target variability to improve the effectiveness of the model to learn under noisy labels.
Our empirical results suggest that the use of noise in a collaborative learning framework is a promising direction and warrants further investigation.

{\small
\bibliographystyle{ieee_fullname}
\bibliography{egbib}

\begin{thebibliography}{10}\itemsep=-1pt

\bibitem{an1996effects}
Guozhong An.
\newblock The effects of adding noise during backpropagation training on a
  generalization performance.
\newblock {\em Neural computation}, 8(3):643--674, 1996.

\bibitem{arpit2017closer}
Devansh Arpit, Stanis{\l}aw Jastrz{\k{e}}bski, Nicolas Ballas, David Krueger,
  Emmanuel Bengio, Maxinder~S Kanwal, Tegan Maharaj, Asja Fischer, Aaron
  Courville, Yoshua Bengio, et~al.
\newblock A closer look at memorization in deep networks.
\newblock In {\em Proceedings of the 34th International Conference on Machine
  Learning-Volume 70}, pages 233--242. JMLR. org, 2017.

\bibitem{athalye2018obfuscated}
Anish Athalye, Nicholas Carlini, and David Wagner.
\newblock Obfuscated gradients give a false sense of security: Circumventing
  defenses to adversarial examples.
\newblock {\em arXiv preprint arXiv:1802.00420}, 2018.

\bibitem{biggio2013evasion}
Battista Biggio, Igino Corona, Davide Maiorca, Blaine Nelson, Nedim
  {\v{S}}rndi{\'c}, Pavel Laskov, Giorgio Giacinto, and Fabio Roli.
\newblock Evasion attacks against machine learning at test time.
\newblock In {\em Joint European conference on machine learning and knowledge
  discovery in databases}, pages 387--402. Springer, 2013.

\bibitem{blundell2015weight}
Charles Blundell, Julien Cornebise, Koray Kavukcuoglu, and Daan Wierstra.
\newblock Weight uncertainty in neural networks.
\newblock {\em arXiv preprint arXiv:1505.05424}, 2015.

\bibitem{bottou1991stochastic}
L{\'e}on Bottou.
\newblock Stochastic gradient learning in neural networks.
\newblock {\em Proceedings of Neuro-N{\i}mes}, 91(8):12, 1991.

\bibitem{bulo2016dropout}
Samuel~Rota Bul{\`o}, Lorenzo Porzi, and Peter Kontschieder.
\newblock Dropout distillation.
\newblock In {\em International Conference on Machine Learning}, pages 99--107,
  2016.

\bibitem{carlini2017towards}
Nicholas Carlini and David Wagner.
\newblock Towards evaluating the robustness of neural networks.
\newblock In {\em 2017 IEEE Symposium on Security and Privacy (SP)}, pages
  39--57. IEEE, 2017.

\bibitem{cohen2019certified}
Jeremy~M Cohen, Elan Rosenfeld, and J~Zico Kolter.
\newblock Certified adversarial robustness via randomized smoothing.
\newblock {\em arXiv preprint arXiv:1902.02918}, 2019.

\bibitem{darlow2018cinic}
Luke~N Darlow, Elliot~J Crowley, Antreas Antoniou, and Amos~J Storkey.
\newblock Cinic-10 is not imagenet or cifar-10.
\newblock {\em arXiv preprint arXiv:1810.03505}, 2018.

\bibitem{deng2009imagenet}
Jia Deng, Wei Dong, Richard Socher, Li-Jia Li, Kai Li, and Li Fei-Fei.
\newblock Imagenet: A large-scale hierarchical image database.
\newblock In {\em 2009 IEEE conference on computer vision and pattern
  recognition}, pages 248--255. Ieee, 2009.

\bibitem{dhillon2018stochastic}
Guneet~S Dhillon, Kamyar Azizzadenesheli, Zachary~C Lipton, Jeremy Bernstein,
  Jean Kossaifi, Aran Khanna, and Anima Anandkumar.
\newblock Stochastic activation pruning for robust adversarial defense.
\newblock {\em arXiv preprint arXiv:1803.01442}, 2018.

\bibitem{faisal2008noise}
A~Aldo Faisal, Luc~PJ Selen, and Daniel~M Wolpert.
\newblock Noise in the nervous system.
\newblock {\em Nature reviews neuroscience}, 9(4):292, 2008.

\bibitem{frenay2013classification}
Beno{\^\i}t Fr{\'e}nay and Michel Verleysen.
\newblock Classification in the presence of label noise: a survey.
\newblock {\em IEEE transactions on neural networks and learning systems},
  25(5):845--869, 2013.

\bibitem{gal2015dropout}
Y Gal and Z Ghahramani.
\newblock Dropout as a bayesian approximation: Representing model uncertainty
  in deep learning. arxiv, 2015.

\bibitem{goodfellow2014explaining}
Ian~J Goodfellow, Jonathon Shlens, and Christian Szegedy.
\newblock Explaining and harnessing adversarial examples.
\newblock {\em arXiv preprint arXiv:1412.6572}, 2014.

\bibitem{graves2011practical}
Alex Graves.
\newblock Practical variational inference for neural networks.
\newblock In {\em Advances in neural information processing systems}, pages
  2348--2356, 2011.

\bibitem{gu2019using}
Keren Gu, Brandon Yang, Jiquan Ngiam, Quoc Le, and Jonathan Shlens.
\newblock Using videos to evaluate image model robustness.
\newblock {\em arXiv preprint arXiv:1904.10076}, 2019.

\bibitem{gurau2018dropout}
Corina Gurau, Alex Bewley, and Ingmar Posner.
\newblock Dropout distillation for efficiently estimating model confidence.
\newblock {\em arXiv preprint arXiv:1809.10562}, 2018.

\bibitem{he2015deep}
K He, X Zhang, S Ren, and J Sun.
\newblock Deep residual learning for image recognition. computer vision and
  pattern recognition (cvpr).
\newblock In {\em 2016 IEEE Conference on}, volume~5, page~6, 2015.

\bibitem{hendrycks2019benchmarking}
Dan Hendrycks and Thomas Dietterich.
\newblock Benchmarking neural network robustness to common corruptions and
  perturbations.
\newblock {\em arXiv preprint arXiv:1903.12261}, 2019.

\bibitem{hendrycks2019natural}
Dan Hendrycks, Kevin Zhao, Steven Basart, Jacob Steinhardt, and Dawn Song.
\newblock Natural adversarial examples.
\newblock {\em arXiv preprint arXiv:1907.07174}, 2019.

\bibitem{hinton2015distilling}
Geoffrey Hinton, Oriol Vinyals, and Jeff Dean.
\newblock Distilling the knowledge in a neural network.
\newblock {\em arXiv preprint arXiv:1503.02531}, 2015.

\bibitem{ilyas2019adversarial}
Andrew Ilyas, Shibani Santurkar, Dimitris Tsipras, Logan Engstrom, Brandon
  Tran, and Aleksander Madry.
\newblock Adversarial examples are not bugs, they are features.
\newblock {\em arXiv preprint arXiv:1905.02175}, 2019.

\bibitem{kawaguchi2017generalization}
Kenji Kawaguchi, Leslie~Pack Kaelbling, and Yoshua Bengio.
\newblock Generalization in deep learning.
\newblock {\em arXiv preprint arXiv:1710.05468}, 2017.

\bibitem{kleinberg2018alternative}
Robert Kleinberg, Yuanzhi Li, and Yang Yuan.
\newblock An alternative view: When does sgd escape local minima?
\newblock {\em arXiv preprint arXiv:1802.06175}, 2018.

\bibitem{korteling2018neural}
Johan~E Korteling, Anne-Marie Brouwer, and Alexander Toet.
\newblock A neural network framework for cognitive bias.
\newblock {\em Frontiers in psychology}, 9, 2018.

\bibitem{krizhevsky2009learning}
Alex Krizhevsky, Geoffrey Hinton, et~al.
\newblock Learning multiple layers of features from tiny images.
\newblock Technical report, Citeseer, 2009.

\bibitem{kurakin2016adversarial}
Alexey Kurakin, Ian Goodfellow, and Samy Bengio.
\newblock Adversarial examples in the physical world.
\newblock {\em arXiv preprint arXiv:1607.02533}, 2016.

\bibitem{lamb2019interpolated}
Alex Lamb, Vikas Verma, Juho Kannala, and Yoshua Bengio.
\newblock Interpolated adversarial training: Achieving robust neural networks
  without sacrificing accuracy.
\newblock {\em arXiv preprint arXiv:1906.06784}, 2019.

\bibitem{lecuyer2018certified}
Mathias Lecuyer, Vaggelis Atlidakis, Roxana Geambasu, Daniel Hsu, and Suman
  Jana.
\newblock Certified robustness to adversarial examples with differential
  privacy.
\newblock {\em arXiv preprint arXiv:1802.03471}, 2018.

\bibitem{li2017convergence}
Yuanzhi Li and Yang Yuan.
\newblock Convergence analysis of two-layer neural networks with relu
  activation.
\newblock In {\em Advances in Neural Information Processing Systems}, pages
  597--607, 2017.

\bibitem{liang2017enhancing}
Shiyu Liang, Yixuan Li, and R Srikant.
\newblock Enhancing the reliability of out-of-distribution image detection in
  neural networks.
\newblock {\em arXiv preprint arXiv:1706.02690}, 2017.

\bibitem{liu2018towards}
Xuanqing Liu, Minhao Cheng, Huan Zhang, and Cho-Jui Hsieh.
\newblock Towards robust neural networks via random self-ensemble.
\newblock In {\em Proceedings of the European Conference on Computer Vision
  (ECCV)}, pages 369--385, 2018.

\bibitem{maass2014noise}
Wolfgang Maass.
\newblock Noise as a resource for computation and learning in networks of
  spiking neurons.
\newblock {\em Proceedings of the IEEE}, 102(5):860--880, 2014.

\bibitem{madry2017towards}
Aleksander Madry, Aleksandar Makelov, Ludwig Schmidt, Dimitris Tsipras, and
  Adrian Vladu.
\newblock Towards deep learning models resistant to adversarial attacks.
\newblock {\em arXiv preprint arXiv:1706.06083}, 2017.

\bibitem{mcdonnell2011benefits}
Mark~D McDonnell and Lawrence~M Ward.
\newblock The benefits of noise in neural systems: bridging theory and
  experiment.
\newblock {\em Nature Reviews Neuroscience}, 12(7):415--425, 2011.

\bibitem{moosavi2016deepfool}
Seyed-Mohsen Moosavi-Dezfooli, Alhussein Fawzi, and Pascal Frossard.
\newblock Deepfool: a simple and accurate method to fool deep neural networks.
\newblock In {\em Proceedings of the IEEE conference on computer vision and
  pattern recognition}, pages 2574--2582, 2016.

\bibitem{moreno2012unifying}
Jose~G Moreno-Torres, Troy Raeder, Roc{\'\i}O Alaiz-Rodr{\'\i}Guez, Nitesh~V
  Chawla, and Francisco Herrera.
\newblock A unifying view on dataset shift in classification.
\newblock {\em Pattern Recognition}, 45(1):521--530, 2012.

\bibitem{muller2019does}
Rafael M{\"u}ller, Simon Kornblith, and Geoffrey Hinton.
\newblock When does label smoothing help?
\newblock {\em arXiv preprint arXiv:1906.02629}, 2019.

\bibitem{neelakantan2015adding}
Arvind Neelakantan, Luke Vilnis, Quoc~V Le, Ilya Sutskever, Lukasz Kaiser,
  Karol Kurach, and James Martens.
\newblock Adding gradient noise improves learning for very deep networks.
\newblock {\em arXiv preprint arXiv:1511.06807}, 2015.

\bibitem{netzer2011reading}
Yuval Netzer, Tao Wang, Adam Coates, Alessandro Bissacco, Bo Wu, and Andrew~Y
  Ng.
\newblock Reading digits in natural images with unsupervised feature learning.
\newblock {\em In NeurIPS workshop on deep learning and unsupervised feature
  learning}, 2011.

\bibitem{park2019relational}
Wonpyo Park, Dongju Kim, Yan Lu, and Minsu Cho.
\newblock Relational knowledge distillation.
\newblock In {\em Proceedings of the IEEE Conference on Computer Vision and
  Pattern Recognition}, pages 3967--3976, 2019.

\bibitem{pereyra2017regularizing}
Gabriel Pereyra, George Tucker, Jan Chorowski, {\L}ukasz Kaiser, and Geoffrey
  Hinton.
\newblock Regularizing neural networks by penalizing confident output
  distributions.
\newblock {\em arXiv preprint arXiv:1701.06548}, 2017.

\bibitem{pinot2019theoretical}
Rafael Pinot, Laurent Meunier, Alexandre Araujo, Hisashi Kashima, Florian Yger,
  C{\'e}dric Gouy-Pailler, and Jamal Atif.
\newblock Theoretical evidence for adversarial robustness through
  randomization: the case of the exponential family.
\newblock {\em arXiv preprint arXiv:1902.01148}, 2019.

\bibitem{rakin2018parametric}
Adnan~Siraj Rakin, Zhezhi He, and Deliang Fan.
\newblock Parametric noise injection: Trainable randomness to improve deep
  neural network robustness against adversarial attack.
\newblock {\em arXiv preprint arXiv:1811.09310}, 2018.

\bibitem{sarfraz2020knowledge}
Fahad Sarfraz, Elahe Arani, and Bahram Zonooz.
\newblock Knowledge distillation beyond model compression.
\newblock {\em arXiv preprint arXiv:2007.01922}, 2020.

\bibitem{scaglione2011trial}
Alessandro Scaglione, Karen~A Moxon, Juan Aguilar, and Guglielmo Foffani.
\newblock Trial-to-trial variability in the responses of neurons carries
  information about stimulus location in the rat whisker thalamus.
\newblock {\em Proceedings of the National Academy of Sciences},
  108(36):14956--14961, 2011.

\bibitem{shimodaira2000improving}
Hidetoshi Shimodaira.
\newblock Improving predictive inference under covariate shift by weighting the
  log-likelihood function.
\newblock {\em Journal of statistical planning and inference}, 90(2):227--244,
  2000.

\bibitem{srivastava2014dropout}
Nitish Srivastava, Geoffrey Hinton, Alex Krizhevsky, Ilya Sutskever, and Ruslan
  Salakhutdinov.
\newblock Dropout: a simple way to prevent neural networks from overfitting.
\newblock {\em The journal of machine learning research}, 15(1):1929--1958,
  2014.

\bibitem{steijvers1996recurrent}
Mark Steijvers and Peter Gr{\"u}nwald.
\newblock A recurrent network that performs a context-sensitive prediction
  task.
\newblock In {\em Proceedings of the 18th annual conference of the cognitive
  science society}, pages 335--339, 1996.

\bibitem{szegedy2016rethinking}
Christian Szegedy, Vincent Vanhoucke, Sergey Ioffe, Jon Shlens, and Zbigniew
  Wojna.
\newblock Rethinking the inception architecture for computer vision.
\newblock In {\em Proceedings of the IEEE conference on computer vision and
  pattern recognition}, pages 2818--2826, 2016.

\bibitem{szegedy2013intriguing}
Christian Szegedy, Wojciech Zaremba, Ilya Sutskever, Joan Bruna, Dumitru Erhan,
  Ian Goodfellow, and Rob Fergus.
\newblock Intriguing properties of neural networks.
\newblock {\em arXiv preprint arXiv:1312.6199}, 2013.

\bibitem{tsipras2018robustness}
Dimitris Tsipras, Shibani Santurkar, Logan Engstrom, Alexander Turner, and
  Aleksander Madry.
\newblock Robustness may be at odds with accuracy.
\newblock {\em arXiv preprint arXiv:1805.12152}, 2018.

\bibitem{tung2019similarity}
Frederick Tung and Greg Mori.
\newblock Similarity-preserving knowledge distillation.
\newblock {\em arXiv preprint arXiv:1907.09682}, 2019.

\bibitem{wan2013regularization}
Li Wan, Matthew Zeiler, Sixin Zhang, Yann Le~Cun, and Rob Fergus.
\newblock Regularization of neural networks using dropconnect.
\newblock In {\em International conference on machine learning}, pages
  1058--1066, 2013.

\bibitem{xie2017mitigating}
Cihang Xie, Jianyu Wang, Zhishuai Zhang, Zhou Ren, and Alan Yuille.
\newblock Mitigating adversarial effects through randomization.
\newblock {\em arXiv preprint arXiv:1711.01991}, 2017.

\bibitem{yim2017gift}
Junho Yim, Donggyu Joo, Jihoon Bae, and Junmo Kim.
\newblock A gift from knowledge distillation: Fast optimization, network
  minimization and transfer learning.
\newblock In {\em Proceedings of the IEEE Conference on Computer Vision and
  Pattern Recognition}, pages 4133--4141, 2017.

\bibitem{zagoruyko2016paying}
Sergey Zagoruyko and Nikos Komodakis.
\newblock Paying more attention to attention: Improving the performance of
  convolutional neural networks via attention transfer.
\newblock {\em arXiv preprint arXiv:1612.03928}, 2016.

\bibitem{zagoruyko2016wide}
Sergey Zagoruyko and Nikos Komodakis.
\newblock Wide residual networks.
\newblock {\em arXiv preprint arXiv:1605.07146}, 2016.

\bibitem{zhang2019theoretically}
Hongyang Zhang, Yaodong Yu, Jiantao Jiao, Eric~P Xing, Laurent~El Ghaoui, and
  Michael~I Jordan.
\newblock Theoretically principled trade-off between robustness and accuracy.
\newblock {\em arXiv preprint arXiv:1901.08573}, 2019.

\bibitem{zhou2019towards}
Mo Zhou, Tianyi Liu, Yan Li, Dachao Lin, Enlu Zhou, and Tuo Zhao.
\newblock Towards understanding the importance of noise in training neural
  networks.
\newblock {\em arXiv preprint arXiv:1909.03172}, 2019.

\bibitem{zhou2017stochastic}
Zhengyuan Zhou, Panayotis Mertikopoulos, Nicholas Bambos, Stephen Boyd, and
  Peter~W Glynn.
\newblock Stochastic mirror descent in variationally coherent optimization
  problems.
\newblock In {\em Advances in Neural Information Processing Systems}, pages
  7040--7049, 2017.

\end{thebibliography}
}


\appendix
\section{Appendix}
\begin{table}[tbh]
\caption{PGD-20 robustness evaluation for Soft Randomization (SR) vs Gaussian Augmentation (GA). GA and SR indicate Gaussian Augmentation and Soft Randomization, respectively. We report the mean and 1 std of the worst-case PGD-20 attack with five random initializations for five different seeds.}
\label{tab:soft_rand_vals}
\centering
\begin{tabular}{|c|cc|cc|}
\hline
 & \multicolumn{2}{c|}{CIFAR-10} & \multicolumn{2}{c|}{SVHN} \\ \cline{2-5}
$\sigma$ & GA & SR & GA & SR \\ \hline
0.01 & 0.00$\pm$0.00 & \textbf{0.78$\pm$0.20} & 0.08$\pm$0.02 & \textbf{21.07$\pm$0.97} \\
0.02 & 0.00$\pm$0.00 & \textbf{3.83$\pm$0.28} & 0.21$\pm$0.06 & \textbf{27.82$\pm$0.83} \\
0.03 & 0.02$\pm$0.01 & \textbf{8.30$\pm$0.79} & 0.52$\pm$0.08 & \textbf{33.41$\pm$0.71} \\
0.04 & 0.12$\pm$0.03 & \textbf{12.89$\pm$0.70} & 1.09$\pm$0.23 & \textbf{39.97$\pm$1.50} \\
0.05 & 0.38$\pm$0.04 & \textbf{15.56$\pm$0.44} & 1.99$\pm$0.13 & \textbf{41.75$\pm$1.06} \\
0.1 & 4.20$\pm$0.38 & \textbf{22.15$\pm$0.40} & 10.82$\pm$1.46 & \textbf{50.35$\pm$1.45} \\
0.2 & 12.19$\pm$0.29 & \textbf{24.14$\pm$0.63} & 25.94$\pm$1.05 & \textbf{51.39$\pm$0.80} \\
0.3 & 17.64$\pm$0.24 & \textbf{23.29$\pm$0.23} & 30.97$\pm$0.19 & \textbf{45.99$\pm$0.32} \\
0.4 & 20.98$\pm$0.16 & \textbf{23.42$\pm$0.32} & 34.34$\pm$0.32 & \textbf{41.28$\pm$0.38} \\
0.5 & 22.34$\pm$0.33 & \textbf{23.54$\pm$0.37} & 37.78$\pm$0.60 & \textbf{40.13$\pm$0.61} \\ \hline
\end{tabular}
\end{table}

\begin{table}[tbh]
\caption {Generalization performance of Messy Collaboration on CIFAR-10 with varying noise rates trained on corrupted labels. We report the mean and 1 std for five different seeds.}
\label{tab:soft_rand_cifar}
\centering
\begin{tabular}{|c|cccccc|}
\hline
Noise rate (r) & 0 & 0.05 & 0.10 & 0.15 & 0.25 & 0.50 \\ \hline
Teacher & 95.11$\pm$0.00 & 92.61$\pm$0.00 & 89.37$\pm$0.00 & 86.56$\pm$0.00 & 78.94$\pm$0.00 & 55.53$\pm$0.00 \\
Baseline & 93.95$\pm$0.18 & 90.51$\pm$0.10 & 86.91$\pm$0.78 & 83.51$\pm$0.47 & 76.07$\pm$1.03 & 56.61$\pm$1.16 \\ \hline \hline
0 & 94.26$\pm$0.16 & \textbf{92.67$\pm$0.14} & 90.69$\pm$0.13 & 89.36$\pm$0.47 & 85.32$\pm$0.52 & 67.97$\pm$0.40 \\
0.05 & 94.39$\pm$0.12 & 92.57$\pm$0.19 & 90.88$\pm$0.25 & 89.36$\pm$0.26 & 85.30$\pm$0.20 & \textbf{68.75$\pm$0.78} \\
0.1 & \textbf{94.41$\pm$0.21} & 92.53$\pm$0.13 & 90.73$\pm$0.26 & 89.33$\pm$0.25 & 85.22$\pm$0.03 & 68.25$\pm$0.50 \\
0.15 & 94.18$\pm$0.08 & 92.52$\pm$0.09 & 90.95$\pm$0.10 & \textbf{89.54$\pm$0.17} & 85.15$\pm$0.28 & 68.27$\pm$0.99 \\
0.25 & 94.32$\pm$0.13 & 92.60$\pm$0.24 & \textbf{90.97$\pm$0.12} & 89.44$\pm$0.24 & 84.96$\pm$0.24 & 68.05$\pm$0.56 \\
0.5 & 94.40$\pm$0.19 & 92.62$\pm$0.20 & 90.96$\pm$0.18 & 89.47$\pm$0.36 & \textbf{85.46$\pm$0.36} & 68.54$\pm$0.55 \\ \hline
\end{tabular}
\end{table}

\begin{table*}[t!h]
\caption {Generalization performance of Messy Collaboration on SVHN with varying noise rates trained on corrupted labels. We report the mean and 1 std for five different seeds.}
\label{tab:soft_rand_svhn}
\centering
\begin{tabular}{|c|cccccc|}
\hline
Noise rate (r) & 0 & 0.05 & 0.10 & 0.15 & 0.25 & 0.50 \\ \hline
Teacher & 96.23$\pm$0.00 & 94.07$\pm$0.00 & 93.19$\pm$0.00 & 91.82$\pm$0.00 & 89.53$\pm$0.00 & 82.84$\pm$0.00 \\
Baseline & 96.11$\pm$0.10 & 93.92$\pm$0.18 & 92.78$\pm$0.38 & 91.58$\pm$0.18 & 89.01$\pm$0.59 & 81.56$\pm$1.13 \\ \hline \hline
0 & \textbf{96.81$\pm$0.07} & 96.29$\pm$0.13 & 95.94$\pm$0.12 & 95.61$\pm$0.14 & \textbf{94.80$\pm$0.09} & \textbf{91.93$\pm$0.25} \\
0.05 & 96.77$\pm$0.16 & 96.30$\pm$0.05 & 95.98$\pm$0.11 & 95.61$\pm$0.08 & 94.79$\pm$0.13 & 91.74$\pm$0.24 \\
0.1 & 96.72$\pm$0.14 & \textbf{96.34$\pm$0.09} & 95.95$\pm$0.09 & 95.63$\pm$0.13 & 94.73$\pm$0.11 & 91.74$\pm$0.12 \\
0.15 & 96.80$\pm$0.13 & 96.31$\pm$0.07 & \textbf{95.99$\pm$0.11} & 95.65$\pm$0.12 & 94.77$\pm$0.17 & 91.73$\pm$0.09 \\
0.25 & 96.77$\pm$0.03 & 96.31$\pm$0.07 & 95.93$\pm$0.14 & \textbf{95.66$\pm$0.13} & 94.67$\pm$0.15 & 91.52$\pm$0.12 \\
0.5 & 96.81$\pm$0.13 & 96.27$\pm$0.07 & 95.90$\pm$0.08 & 95.59$\pm$0.08 & 94.59$\pm$0.10 & 91.05$\pm$0.13 \\ \hline
\end{tabular}
\end{table*}

\begin{table*}[!ht]
\caption {Out-of-distribution generalization performance on CINIC dataset for models trained with Messy Collaboration on CIFAR-10 with varying noise rates trained on corrupted labels. We report the mean and 1 std for five different seeds.}
\label{tab:soft_rand_cinic}
\centering
\begin{tabular}{|c|cccccc|}
\hline
Teacher & 70.23$\pm$0.00 & 66.22$\pm$0.00 & 62.66$\pm$0.00 & 59.07$\pm$0.00 & 53.63$\pm$0.00 & 36.75$\pm$0.00 \\
Baseline & 68.89$\pm$0.08 & 63.22$\pm$0.53 & 59.50$\pm$0.27 & 55.00$\pm$0.79 & 50.08$\pm$1.03 & 37.21$\pm$0.99 \\ \hline \hline
0 & 68.95$\pm$0.18 & 65.52$\pm$0.34 & \textbf{63.83$\pm$0.14} & 61.24$\pm$0.30 & 56.68$\pm$0.44 & 44.42$\pm$0.73 \\
0.05 & 69.04$\pm$0.13 & 65.52$\pm$0.82 & 63.69$\pm$0.16 & 61.28$\pm$0.48 & 56.58$\pm$0.42 & \textbf{45.15$\pm$0.21} \\
0.1 & 69.02$\pm$0.17 & \textbf{65.97$\pm$0.38} & 63.57$\pm$0.34 & \textbf{61.68$\pm$0.39} & 56.15$\pm$0.62 & 44.92$\pm$0.58 \\
0.15 & \textbf{69.20$\pm$0.15} & 65.74$\pm$0.33 & 63.73$\pm$0.47 & 61.34$\pm$0.47 & 56.41$\pm$0.51 & 44.78$\pm$0.57 \\
0.25 & 69.13$\pm$0.19 & 65.92$\pm$0.34 & 63.45$\pm$0.32 & 61.52$\pm$0.52 & \textbf{56.74$\pm$0.53} & 45.02$\pm$0.56 \\
0.5 & 69.06$\pm$0.21 & 65.28$\pm$0.44 & 63.62$\pm$0.41 & 61.19$\pm$0.19 & 56.40$\pm$0.42 & 44.68$\pm$1.16 \\ \hline 
\end{tabular}
\end{table*}

\end{document}